\documentclass[conference]{IEEEtran}
\IEEEoverridecommandlockouts
\usepackage{cite}
\usepackage{amsmath,amssymb,bm,amsfonts}
\usepackage{algorithmic}
\usepackage{graphicx}
\usepackage{textcomp}
\usepackage{hyperref}
\usepackage{xspace}
\usepackage{caption,subcaption,graphicx}
\usepackage{multirow}
\usepackage{xcolor}
\def\BibTeX{{\rm B\kern-.05em{\sc i\kern-.025em b}\kern-.08em
    T\kern-.1667em\lower.7ex\hbox{E}\kern-.125emX}}

\newcommand{\model}{\texttt{GLAD}}

\begin{document}

\title{GLAD: Content-aware Dynamic Graphs For Log Anomaly Detection}


\author{
Yufei Li$^{1}$
\xspace    
Yanchi Liu$^{2}$
\xspace  
Haoyu Wang$^{2}$
\xspace 
Zhengzhang Chen$^{2}$
\xspace 
Wei Cheng$^{2}$
\xspace 
Yuncong Chen$^{2}$
\xspace 
Wenchao Yu$^{2}$
\xspace
\\
Haifeng Chen$^{2}$
\xspace  
Cong Liu$^{1}$\\
$^{1}$University of California, Riverside\xspace\xspace
$^{2}$NEC Labs America\\
$^{1}$\{yli927,congl\}@ucr.edu\xspace\xspace 
$^{2}$\{yanchi,haoyu,zchen,weicheng,yuncong,wenchao,haifeng\}@nec-labs.com}

\maketitle

\begin{abstract}

Logs play a crucial role in system monitoring and debugging by recording valuable system information, including events and states. 
Although various methods have been proposed to detect anomalies in log sequences, they often overlook the significance of considering relations among system components, such as services and users, which can be identified from log contents. 
Understanding these relations is vital for detecting anomalies and their underlying causes. 
To address this issue, we introduce \model{}, a Graph-based Log Anomaly Detection framework designed to detect relational anomalies in system logs. 
\model{} incorporates log semantics, relational patterns, and sequential patterns into a unified framework for anomaly detection.
Specifically, \model{} first introduces a field extraction module that utilizes prompt-based few-shot learning to identify essential fields from log contents. 
Then \model{} constructs dynamic log graphs for sliding windows by interconnecting extracted fields and log events parsed from the log parser. 
These graphs represent events and fields as nodes and their relations as edges. 
Subsequently, \model{} utilizes a temporal-attentive graph edge anomaly detection model for identifying anomalous relations in these dynamic log graphs. 
This model employs a Graph Neural Network (GNN)-based encoder enhanced with transformers to capture content, structural and temporal features.
We evaluate our proposed method\footnote{Our code is available at \url{https://github.com/yul091/GraphLogAD}} on three datasets, and the results demonstrate the effectiveness of \model{} in detecting anomalies indicated by varying relational patterns.
\end{abstract}


\section{Introduction}

Anomaly detection is the task of identifying unusual or unexpected behaviors in a system or process.
As computer systems become increasingly more sophisticated due to the expansion of new communication technologies and services, they are prone to various adversarial attacks and bugs~\cite{log-anomaly-detection-survey}.
Moreover, such attacks are also getting evolved and becoming increasingly sophisticated.
As a result, the difficulty of anomaly detection has increased, making many conventional detection approaches no longer effective, and it requires us to look deeper into the system, for example, the interaction among system components.

\begin{figure}[]
    \centering
    \includegraphics[width=0.4\textwidth]{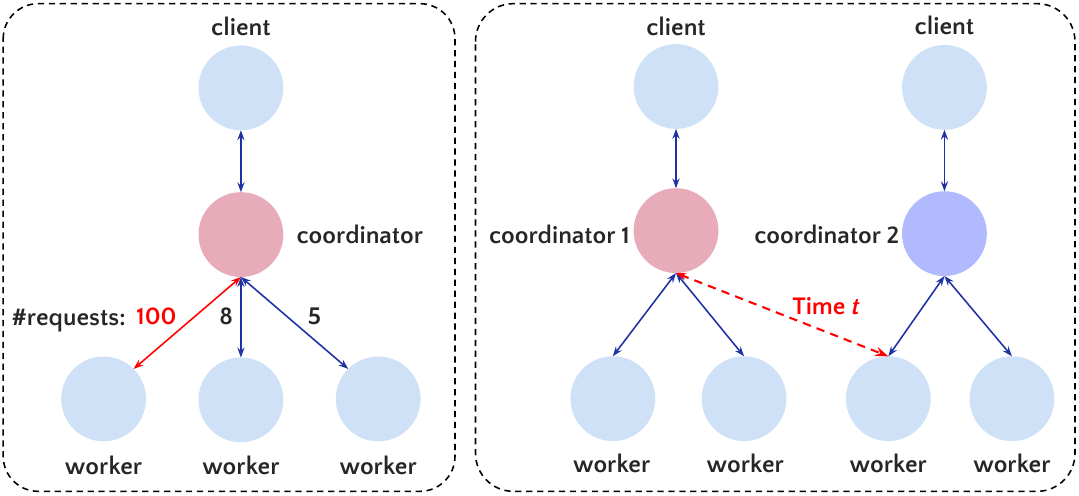}
    \caption{Two anomalous relations (anomalous edges highlighted in \textcolor{red}{red}): unbalanced request (left) and malicious request (right).}
    \label{fig:anomaly_example}
    \vspace{-0.3cm}
\end{figure}

System logs capture system states and events across time to aid process monitoring and root cause analysis of running services.
These log files are ubiquitous in almost all computer systems and contain rich information, including control commands of machine systems, transactions of customer purchases, and logs of a computer program.
As a result, they have proven a valuable resource for anomaly detection in both academic research and industry applications~\cite{AD-applications-airline-safety,anomaly-detection-survey,event-sequence-AD,efficient-AD-security,ccs-2017-deeplog}.
Each log message usually consists of a predefined constant key template (known as a ``event'', e.g., a login activity) and a few variables (known as ``fields'', e.g, \textit{services} and \textit{users}).
When the events are arranged chronologically based on the recording time, they form a discrete log sequence. Various methods have been proposed to detect the anomalous sequential patterns in the sequence:
(1) \textit{Pattern recognition} methods consider event sequences with inconsistencies beyond a certain threshold to be anomalous~\cite{high-dimension-distribution,log-clustering,isolation-forest-2008,mining-invariants-log,large-scale-mining-console-logs}.
They treat event alphabet sequence as input in an independent dimension and ignore the sequential patterns between events.
(2) \textit{Sequential learning} methods analyze events sequentially with a defined sliding window in order to forecast the subsequent event based on the observation window~\cite{ccs-2017-deeplog,logbert}.

However, the relation between log events and fields, an essential indicator of system anomalies, has often been overlooked.
This oversight can lead to missed detection or false alarms, as anomalies may not be apparent from individual events or isolated patterns.
Different from previous methods that detect anomalous sequential patterns in log sequences, we focus on a new task that aims at detecting anomalous relational patterns between interconnected events and fields. Take, for instance, a scenario where workers receive an unbalanced number of requests from a coordinator in a period of time, or a coordinator suddenly requests connection to other workers, as illustrated in Figure~\ref{fig:anomaly_example}.
Traditional methods, without considering the relations, may fall short in detecting such anomalies. 
Apart from detecting anomalous events, understanding these anomalous relations between events can offer insightful details about the system's dynamics, for example, the underlying causes of an anomaly and its propagation over time.

To achieve our goal, there are several challenges: 
(1) Dynamic graphs need to be built to describe the interactions between log events and fields in different time windows. 
(2) Instead of merely detecting anomalies on graph level, we aim to detect anomalous edges representing the relations among nodes, which is a more challenging task.
(3) In addition to relational patterns, we need to integrate log semantics and sequential patterns as a whole for anomaly detection.
To this end, we propose \model{}, a \textbf{G}raph-based \textbf{L}og \textbf{A}nomaly \textbf{D}etection framework, to extract and learn the relations among log events and fields, in addition to log semantics and sequential patterns, for system relation anomaly detection.
Our approach proposes a novel method to construct dynamic graphs that describe the relations among log events and fields over time and then leverages a temporal-attentive transformer to capture the sequential patterns implicitly expressed in each time period. 
Specifically, a field extraction module utilizing prompt-based few-shot learning is first used to extract field information from log contents.
Then, with the fields extracted and the log events parsed from a log parser, dynamic graphs can be constructed for sliding windows with events and fields as nodes and the relations between them as edges. 
Finally, a temporal-attentive graph edge anomaly detection method is proposed to detect anomalous relations from evolving graphs, where a Graph Neural Network (GNN)-based encoder facilitated with transformers is used to learn the structural, content, and sequential features. 
Experiments on real-world log datasets are conducted to demonstrate the effectiveness of \model{}.

To summarize, in this work, we propose to detect log anomalies from a novel point of view, i.e., the interaction and relation between system components leveraging system logs.
In this way, we can dig into more system details and find causes and solutions to the anomalies efficiently. 
Our main contribution is a framework for constructing dynamic graphs from logs and capturing relational anomalies from dynamic graphs using temporal-attentive transformers, which allows for more granular and accurate log anomaly detection. 
We believe our proposed approach has the potential to significantly improve the effectiveness of log analysis in detecting more sophisticated anomalies in real applications.

\section{Related Work}

\begin{table}[]
\caption{Notation Description.}
    \centering
    \resizebox{0.45\textwidth}{!}{
    \begin{tabular}{cl}
    \hline
    Symbol & Description \\
    \hline
    \hline
    $e$ & $e=\left \{ x_{1}, ..., x_{|e|} \right \}$ log message is a sequence of tokens \\
    $S$ & $S=\left \{ e_{1}, ..., e_{|S|} \right \}$ log sequence is a sequential series of logs \\
    $E$ & $E=\left \{ ent_1,...,ent_{|E|} \right \}$ sequence of entities in a log message \\
    $Y$ & $Y=\left \{ l_1,...,l_{|E|} \right \}$ sequence of entity labels in a log message \\
    $\mathcal{S}$ & $\mathcal{S}=\left \{ S_{1}, ..., S_{|\mathcal{S}|} \right \}$ total sequences are a set of log sequences \\
    \hline
    $\mathcal{G}_t$ & the dynamic graph at time window $t$ with $\mathcal{V}_t$ and $\mathcal{E}_t$ \\
    $\mathcal{V}_t$ & vertex set in graph $\mathcal{G}_t$ \\
    $\mathcal{E}_t$ & edge set in graph $\mathcal{G}_t$ \\
    $\mathbf{X}_t$ & attribute matrix in graph $\mathcal{G}_t$ \\
    $\mathbf{A}_t$ & adjacency matrix in graph $\mathcal{G}_t$ \\
    $\mathbf{W}^{(l)}$ & learnable weights in the $l$-th layer of a model, e.g., $\mathbf{W}_{ner}$, $\mathbf{W}_g^{(l)}$ \\
    $\mathbf{I}$ & identity matrix \\
    $\mathbf{H}_{t}$ & node representations of graph $\mathcal{G}_t$ learned by GCN \\
    $N$ & total number of graphs in $\mathcal{S}$ \\
    $\bm{\mathcal{H}}_{\mathcal{S}, t}$ & long-term node representations of graph $\mathcal{G}_t$ learned by transformers \\
    $\bm{\mathcal{H}}_{k, t}$ & short-term node representations of graph $\mathcal{G}_t$ learned by transformers \\
    $\bm{\mathcal{H}}_t$ & node representations of graph $\mathcal{G}_t$ by concatenating $\bm{\mathcal{H}}_{\mathcal{S}, t}$ and $\bm{\mathcal{H}}_{k, t}$ \\
    $\bm{\mathcal{R}}_{t}$ & graph representation of $\mathcal{G}_t$ by maxpooling node representations $\bm{\mathcal{H}}_t$
    \\
    \hline
    $\sigma(\cdot)$ & activation function, e.g., ReLU($\cdot$), Sigmoid($\cdot$) \\
    $\mathcal{L}$ & loss objective, including $\mathcal{L}_{ner}$, $\mathcal{L}^t$, $\mathcal{L}_{reg}$ \\
    $\mathbf{P}$ & prompt $\mathbf{P}=\left \{ p_1,...,p_m \right \}$, including $\mathbf{P}^{+}$ and $\mathbf{P}^{-}$ \\
    \hline
    \end{tabular}}
    \label{tab:terminologies}
    \vspace{-0.3cm}
\end{table}

\noindent\textbf{Log Sequences Anomaly Detection.}
Detecting anomalies in log sequences has recently gained substantial attention.
Earlier research hinged upon similarity measurements, wherein test logs are compared with training logs to detect anomalies based on their dissimilarity~\cite{Similarity-measure,similarity-survey}.
Subsequent methods can be categorized into three groups: \textit{pattern frequency-based}~\cite{pattern-based-anomaly-detection}, \textit{sequence-based} such as Hidden Markov Model~(HMM)~\cite{HMM-based-anomaly-detection}, and \textit{contiguous subsequence-based} anomaly detection such as window-based techniques~\cite{window-based-anomaly-detection, anomaly-detection-survey}.
While certain studies utilize supervised learning for anomaly detection~\cite{classification-AD,logtransfer,failure-prediction,semi-supervised-ad}, unsupervised learning, which observe only normal event sequences during training, has been proven to be a more efficient learning paradigm~\cite{LOF,log-clustering,isolation-forest-2008,mining-invariants-log,large-scale-mining-console-logs,kdd-2022-cat}.
Our research mainly focuses on the latter learning paradigm.

\noindent\textbf{Log Knowledge Graph Construction.}
Raw log files offer a wealth of information pertaining to system states and service interconnection, e.g., whether a computing machine is running under an abnormal state or a user is a malicious attacker.
To analyze such data and avoid tedious searching clues or tracing system events across log sources, existing studies have put efforts into identifying and linking entities (log fields) across log sources, thereby enriching them with knowledge graphs~\cite{LEKG,SLOGERT,ARES}. 
They often apply information extraction techniques such as Named Entity Recognition (NER) to identify log fields within log messages.
The resulting fields are considered nodes within a knowledge graph, and rule-based relation linking is used to integrate the log fields into the knowledge graph.
However, these methods require a large amount of label data for training, which introduce high cost in real applications. In comparison, we try to solve this problem in a few-shot setting.

\noindent\textbf{Graph-based Anomaly Detection.} 
GNNs have become increasingly popular due to their ability to learn relation patterns, making them favorable for anomaly detection.
Leading GNN models include GCN~\cite{GCN}, GIN~\cite{GIN}, SAGE~\cite{SAGE}, GAT~\cite{GAT}, and Transformer Graph~(GT)~\cite{TransformerConv}.
Existing graph-based anomaly detection methods can be categorized into three types based on the range of anomaly detection:
(1) \textit{Node-level auto-encoders}~\cite{dominant,anomalydae,GAAN,conad} regard nodes with atypical attribute and relation distributions as anomalies.
The key idea is to use GNN-based encoder-decoders to reconstruct original graphs and calculate the reconstruction errors for each node.
Nodes with above-threshold errors are detected as anomalies.
Some further consider temporal relations on dynamic graphs~\cite{NetWalk,OCAN} to detect anomalies.
(2) \textit{Edge-level auto-encoders}~\cite{UGED,AANE} first use graph encoders to learn node feature representations, then determine edge scores for each node pair in the graph to represent how likely it is normal. 
Some further consider representative structural information from the dynamic graph in each time stamp and their dependencies ~\cite{NetWalk,structural-temporal-grah, addgraph} to detect anomalous edges. 
(3) \textit{Graph-level auto-encoders}~\cite{DeepSVDD,DeepSphere,FraudNE,DeepFD,UPFD} use a graph encoder to learn feature representations and aggregates all node features within each graph as the graph representation.
Hypersphere learning is then applied to cluster all normal graphs into a central distribution, distinguishing them from anomalous ones.

\section{Log Anomaly Detection Framework}

\begin{figure}
\centering
\includegraphics[width=0.9\columnwidth]{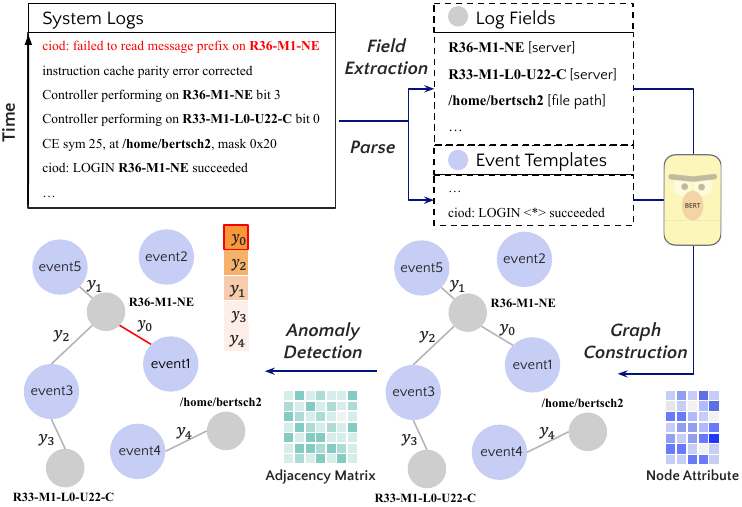} 
\caption{Overview of our \model{} framework. \model{} first extracts log fields and events and connects them to construct dynamic log graphs, where node features are text embeddings. 
These graphs, along with their sequential dependencies, are jointly encoded to identify anomalous edges.}
\label{fig:framework}
\vspace{-0.3cm}
\end{figure}

In this section, we introduce \model{}, a graph-based framework that learns structural, content, and sequential features among logs for anomaly detection, as shown in Figure~\ref{fig:framework}.

\subsection{Preliminaries}
We first define several important terminologies pertinent to our work. 
The notions are summarized in Table~\ref{tab:terminologies}.

A \textbf{log} is a sequence of tokens $e=\left \{ x_{1}, ..., x_{|e|} \right \}$, where $x_i$ denotes the $i$-th token and $|e|$ is the log length.

A \textbf{log sequence} is a series of logs ordered chronologically within an observed time window $S=\left \{ e_{1}, ..., e_{|S|} \right \}$, where $e_i$ represents the $i$-th log and $|S|$ denotes the total number of logs in a time window.

For a log sequence $S_t$ in time window $t$, we construct a \textbf{dynamic graph} $\mathcal{G}_t=(\mathcal{V}_t,\mathcal{E}_t,\mathbf{X}_t,\mathbf{A}_t)$, where $\mathcal{V}_t$, $\mathcal{E}_t$ denote the union of vertices and the union of edges, $\mathbf{X}_t \in \mathbb{R}^{n\times d}$ and $\mathbf{A}_t\in \mathbb{R}^{n\times n}$ are its attribute and adjacency matrices. 
Note that the dynamic graph used in this paper is an undirected, weighted, and attributed heterogeneous graph.

\subsection{Log Graph Construction}
To build graph representations from log sequences, we propose a prompt-based model to extract fields from log messages. 
The extracted fields, along with the parsed log events via a log parser, are interconnected following pre-defined principles to construct dynamic graphs.
Subsequently, we employ a pre-trained Sentence-BERT~\cite{reimers-gurevych-2019-sentence} to capture the semantics of each node using its content information.
The encoded hidden representations for each node are treated as its attributes, while the adjacency matrix represents the structure of the graph. 
These node attributes and adjacency matrices are collectively used to detect anomalous edges.

\noindent\textbf{Prompt-Based Few-Shot Field Extraction.}
Real-world log datasets contain substantial log events and log fields with diverse syntactic formats, making manual annotations virtually infeasible. 
While existing off-the-shelf tools~\cite{DBLP:conf/cikm/HamooniDXZJM16,DBLP:conf/iwpc/MessaoudiPBBS18} employ either rule-based or search-based algorithms to extract event templates and fields from raw log messages, their effectiveness is limited.
They work well with fields exhibiting fixed syntax patterns, such as \textit{IP}, \textit{email}, and \textit{URL}, but falter with those that have flexible syntax patterns, like \textit{user} and \textit{service}.

To overcome this challenge, we approach log field extraction as a NER task and propose a prompt-based few-shot learning method using BART~\cite{BART} that excels in identifying log fields in low-resource scenarios.
We define 15 common log field types vital for system monitoring by referring to common log ontology~\cite{SLOGERT,LEKG,ARES}. 
These include \textit{IP}, \textit{email}, \textit{process ID}~(\textit{pid}), \textit{user ID}~(\textit{uid}), \textit{user name}, \textit{timestamp}, \textit{service}, \textit{server}, \textit{file path}, \textit{URL}, \textit{port}, \textit{session}, \textit{duration}, \textit{domain}, and \textit{version}. 

\begin{figure}[]
\centering
\begin{subfigure}[b]{0.27\textwidth}\includegraphics[width=\textwidth]{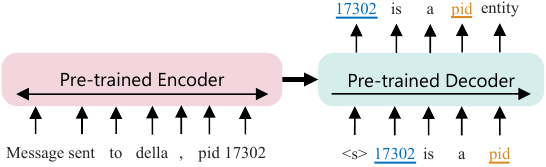}
    \caption{Training of prompt-based method. The prompt used is "$\left \langle x_{i:j} \right \rangle$ is a/an $\left \langle l_k \right \rangle$ entity".} 
    \label{fig:seq2seq_train}
\end{subfigure}
\begin{subfigure}[b]{0.205\textwidth}
\includegraphics[width=\textwidth]{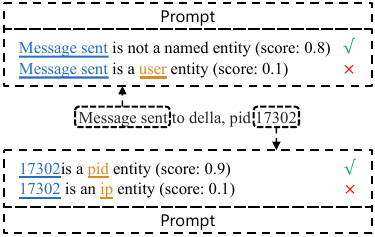}
    \caption{Inference of the prompt-based method.}
    \label{fig:seq2seq_test}
\end{subfigure}
\label{fig:seq2seq}
\caption{Illustration of prompt-based few-shot field extraction.}
\vspace{-0.3cm}
\end{figure}

We frame the field extraction as a seq2seq learning process, as shown in Figure~\ref{fig:seq2seq_train}.
Given a log message $e =\left \{ x_1,...,x_{|e|} \right \}$, which contains a set of gold fields $E=\left \{ ent_1,...,ent_{|E|} \right \}$ and a label set $Y=\left \{ l_1,...,l_{|E|} \right \}$, we create a target sequence (prompt) $\mathbf{P}_{l_k, x_{i:j}} = \left \{ p_1,...,p_m \right \}$ for each candidate text span $x_{i:j}$ and its label $l_k$.
Specifically, $\mathbf{P}$ is a positive prompt $\mathbf{P}^{+}$ if the text span is a gold field ($x_{i:j} \in E$), e.g., ``$\left \langle x_{i:j} \right \rangle$ is a/an $\left \langle l_k \right \rangle$ entity''; otherwise, it is a
negative prompt $\mathbf{P}^{-}$, e.g., ``$\left \langle x_{i:j} \right \rangle$ is not a named entity''.

During training, we create prompts using gold fields following~\cite{DBLP:conf/acl/CuiWLYZ21,li-etal-2023-uncertainty}. 
For each log message $e$, we create positive pairs $(e, \mathbf{P}^{+})$ by traversing all its gold fields and negative pairs $(e, \mathbf{P}^{-})$ by randomly sampling non-entity text spans. 
For efficiency, we limit the number of $n$-grams for a span to 1$\sim$5, i.e., 5$*n$ negative prompts are created for each log message. 
After sampling, the number of negative pairs is three times that of positive pairs. 
Given a sequence pair $(e, \mathbf{P})$, we feed the log message $e$ to the encoder of BART whose hidden size is $d_h$, and obtain the hidden states $\mathbf{h}^{enc}\in \mathbb{R}^{d_h}$:
\begin{equation}
    \mathbf{h}^{enc}=\text{Encoder}(x_{1:|e|})
\end{equation}

At the $c$-th decoding step, $\mathbf{h}^{enc}$ and previous output tokens $p_{1:c-1}$ are used to generate a representation via attention~\cite{DBLP:conf/nips/VaswaniSPUJGKP17}:
\begin{equation}
    \mathbf{h}_c^{dec}=\text{Decoder}(\mathbf{h}^{enc}, p_{1:c-1})
\end{equation}

The conditional probability of a word $p_c$ is defined as:
\begin{equation}
    P(p_c|p_{1:c-1},e)=\text{softmax}(\mathbf{h}_c^{dec}\mathbf{W}_{ner}+\mathbf{b}_{ner})
\end{equation}
where $\mathbf{W}_{ner}\in \mathbb{R}^{d_h\times |V|}$ and $\mathbf{b}_{ner}\in \mathbb{R}^{|V|}$. 
Here $|V|$ denotes the vocab size of BART. 
The decoding objective is the Cross-Entropy~(CE) loss for prompt with length $m$:

\begin{equation}
    \mathcal{L}_{ner}=-\sum_{c=1}^{m}\log{P(p_c|p_{1,c-1}, e)}
\end{equation}

During inference, we enumerate all possible 1$\sim$5-grams text spans $x_{i:j}$ for a log message $e$ and compute scores for each prompt $\mathbf{P}_{l_k,x_{i:j}}=\left \{ p_1,...,p_m \right \}$ as follows:
\begin{equation}
    f(\mathbf{P}_{l_k,x_{i:j}})=\sum_{c=1}^{m}\log{P(p_c|p_{1:c-1}, e)}
\end{equation}

For each traversed text span $x_{i:j}$, we compute the score $f(\mathbf{P}_{l_k,x_{i:j}}^{+})$ for every entity type and $f(\mathbf{P}_{x_{i:j}}^{-})$ for the non-entity type.
A resulting type $l_k^{*}$ than garners the highest score is assigned to $x_{i:j}$.
Such iterative process ensures the extraction of all relevant fields, as depicted in Figure~\ref{fig:seq2seq_test}.

\noindent\textbf{Graph Structure Configuration.}
To model the relation between fields and events across different log messages, we use a sliding window with a fixed time interval to snapshot a batch of log messages and construct a corresponding graph.
Specifically, each log instance consists of a parsed event template (obtained via a log parser such as Drain~\cite{Drain}), e.g., ``FAILED LOGIN for $\left \langle * \right \rangle$ to $\left \langle * \right \rangle$'', along with a list of extracted fields, e.g., [``della'', ``imap://localhost/''] with corresponding types, e.g., [\textit{user}, \textit{server}]. 
We then interconnect the event template to each extracted field to capture inherent behaviors in the log with the number of connections as edge weight.
In the resultant undirected graph, any two log instances that share any of the defined nodes are indirectly connected, thereby indicating their implicit relations.


\noindent\textbf{Graph Node Attribute Configuration.}
We define types of nodes based on the corresponding event and field types, such as \textit{server} for ``imap://localhost/". 
For each node, we define its input text format and employ a pre-trained Sentence-BERT~\cite{reimers-gurevych-2019-sentence} to learn the sentence embedding as its attribute.
Specifically, for log events, we directly use their templates as the encoder input texts, while for log fields we use our defined prompts as the input texts, e.g., ``imap://localhost/ is a server entity''. 
The output hidden states for each input text capture the node semantics and are used as node features for constructing attributed graphs.

\subsection{Temporal-Attentive Graph Edge Anomaly Detection}
\begin{figure}[]
\centering
\includegraphics[width=0.5\textwidth]{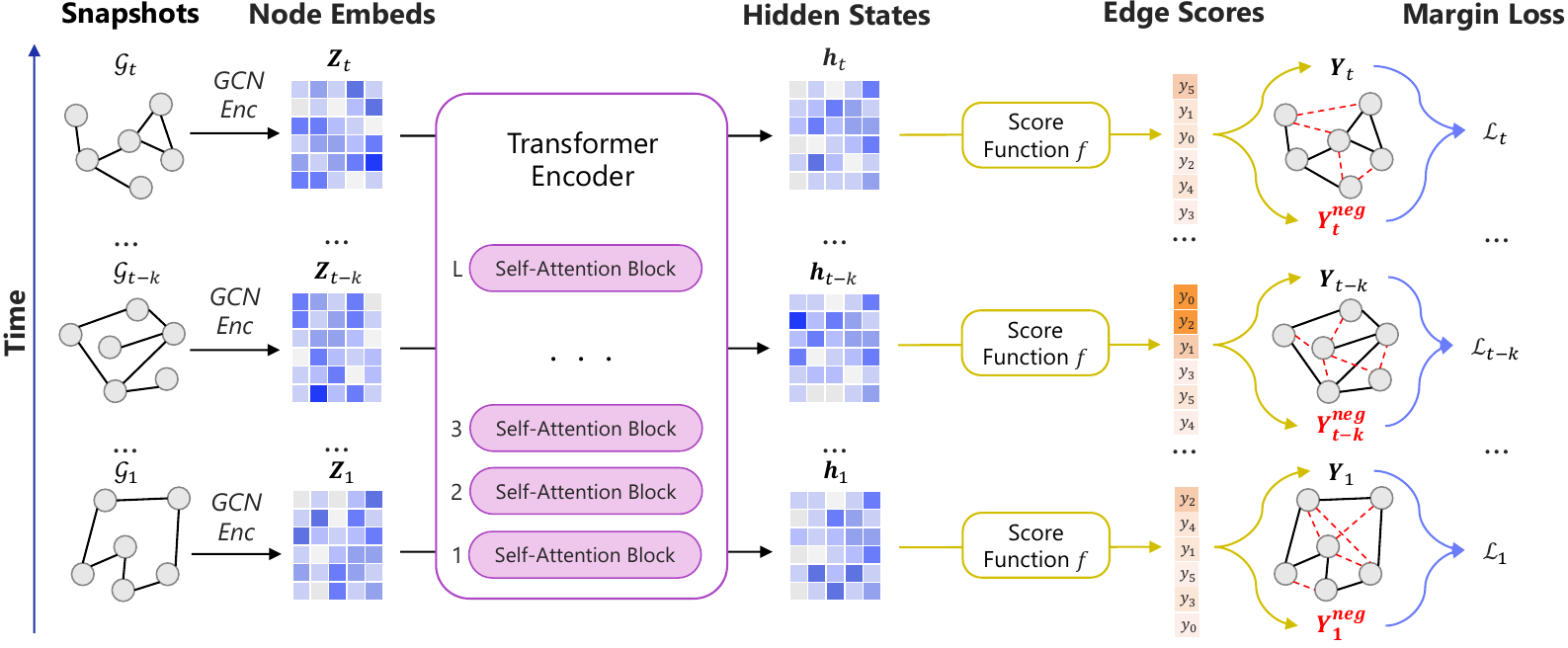}
\caption{Overview of our temporal-attentive graph edge anomaly detection framework. Edges highlighted in \textcolor{red}{red} are negative edges. $\mathbf{Y}_t$ and $\mathbf{Y}_t^{neg}$ denote the aggregation of positive and negative edge scores, respectively, for a specific graph $\mathcal{G}_t$.}
\label{fig:anomaly_detect}
\vspace{-0.3cm}
\end{figure}

We now introduce our proposed temporal-attentive graph edge anomaly detection method, as illustrated in Figure~\ref{fig:anomaly_detect}, which operates on the dynamic graphs constructed for logs within corresponding time slots. 
Specifically, a Graph Convolutional Network~(GCN) is first used to encode the structural information for each graph. 
Then, a transformer encoder is deployed to learn the temporal dependencies within the sequence of dynamic graphs. 
For each graph, we sample certain negative edges and compute the edge score using the learned hidden states. 
The process concludes by employing a pair-wise margin loss to minimize the positive edge scores and maximize the negative edge scores, in adherence with the one-class training objective.

\noindent\textbf{GCN Shared Encoder.}
At time window $t$, we receive a graph snapshot $\mathcal{G}_t=(\mathcal{V}_t,\mathcal{E}_t,\mathbf{X}_t,\mathbf{A}_t)$, where $\mathbf{X}_t \in \mathbb{R}^{n\times d}$ and $\mathbf{A}_t\in \mathbb{R}^{n\times n}$ represent its attribute and adjacency matrices respectively. 
We apply GCN~\cite{GCN} to capture both its attribute and structural features. 
While there exist advanced GNNs, such as Graph Transformer (GT)~\cite{TransformerConv}, we found that GCN offers a blend of efficiency and competitive performance.
It considers high-order node proximity when encoding the embedding representations, thereby alleviating network sparsity beyond the observed links among nodes ~\cite{dominant}. 
For an $L$-layered GCN, each layer can be expressed with the function:
\begin{equation}  
\label{eq:gcn_encoder}
\begin{gathered}
\mathbf{H}^{(l)}_{t} = f_{\sigma}(\mathbf{H}^{(l-1)}_{t}, \mathbf{\hat{A}}_{t}|\mathbf{W}_{g}^{(l)}) \\
f_{\sigma}(\mathbf{H}^{(l)}_{t}, \mathbf{\hat{A}}_{t}|\mathbf{W}_{g}^{(l)})=\sigma(\mathbf{\hat{D}}^{-\frac{1}{2}}_{t} \mathbf{\hat{A}}_{t} \mathbf{\hat{D}}^{-\frac{1}{2}}_{t} \mathbf{H}^{(l)}_{t} \mathbf{W}_{g}^{(l)})
\end{gathered}
\end{equation}
where $\mathbf{W}_{g}^{(l)}$ is a learnable weight matrix for the $l$-th layer, $l\in [1,L]$.
$\mathbf{\hat{A}}_t=\mathbf{A}_t+\mathbf{I}$ denotes the adjacency matrix with added self-loops and $\mathbf{\hat{D}}_{i,i}=\sum_{j=0}\hat{A}_{i,j}$ represents its diagonal degree matrix. 
$\sigma(\cdot)$ is a non-linear activation function, for which we use the ReLU.

We designate the attribute matrix $\mathbf{X}_{t}$ as the initial hidden state $\mathbf{H}^{(0)}_{t}$. 
The resultant embedding $\mathbf{Z}_t  = \mathbf{H}^{(L)}_{t}$ captures the nonlinearity of complex interactions between log entities and events within each graph.
However, it is still inadequate for detecting anomalies caused by malicious relations due to neglect of temporal features across graph snapshots.

\noindent\textbf{Temporal-Attentive Transformer.}
Given the chronologically generated nature of system logs, and the logical dependencies that exist between past and present log states, we employ a transformer encoder to incorporate the temporal features of entire sequence into the latent space.

We receive a sequence of node embeddings $\left \{ \mathbf{Z}_{1}, ..., \mathbf{Z}_{N} \right \}$ for all graphs.
Note that nodes in each graph are an unordered set, $\mathcal{V}_t=\left \{v_{1}, ..., v_{|\mathcal{V}_t|} \right \}$, rather than a sequence.
We propose a Set Transformer~(ST) to eliminate this order dependencies when encoding node embeddings.
Specifically, we first compute the position embeddings based on each graph's position in the sequence, assigning all nodes belonging to each graph the identical position embedding $\mathbf{E}_{p}$.
Subsequently, the embedding for the graph at time $t$ (with position $p$) is determined as $\mathbf{E}_t=\mathbf{E}_{p}+\mathbf{Z}_t$, and the representation sequence as $\mathbf{E}_{\mathcal{S}}=\left \{ \mathbf{E}_{1}, ..., \mathbf{E}_{N} \right \}$.
The representation sequence is then fed into self-attention blocks to derive long-term representations $\bm{\mathcal{H}}_{\mathcal{S}}$:

\begin{equation}
    \bm{\mathcal{H}}_{\mathcal{S}}^{(l+1)}=\text{FFN}(\text{Attention}(\bm{\mathcal{H}}_{\mathcal{S}}^{(l)}))
\end{equation}
where $l$ denotes the layer index, with the initial hidden state $\bm{\mathcal{H}}_{\mathcal{S}}^{(0)}=\mathbf{E}_{\mathcal{S}}$.
We formulate subsequences using a sliding window of size $k$.
Consequently, each subsequence comprises unique local information, pivotal in determining whether the entire sequence is anomalous.
For a subsequence of graph node embeddings $\left \{ \mathbf{Z}_{t-k-1}, ..., \mathbf{Z}_{t} \right \}$, corresponding to graphs $\left \{ \mathcal{G}_{t-k-1}, ..., \mathcal{G}_{t} \right \}$, its representation can be expressed as $\mathbf{E}_{k}=\left \{ \mathbf{E}_{t-k-1}, ..., \mathbf{E}_{t} \right \}$.
The same operations are executed to obtain short-term representations $\bm{\mathcal{H}}_{k}$ by considering $k$ local graphs.

We then concatenate the encoded long-term $\bm{\mathcal{H}}_{\mathcal{S}}$ and short-term representations $\bm{\mathcal{H}}_k$ to form the final node features:
\begin{equation}
    \bm{\mathcal{H}}=[\bm{\mathcal{H}}_{\mathcal{S}}||\bm{\mathcal{H}}_k]_{dim=1}
\end{equation}
where $[\cdot||\cdot]_{dim=1}$ represents the concatenation operator of two matrices over the column-wise dimension.
Consequently, the final node representations $\bm{\mathcal{H}}_t$ for graph $\mathcal{G}_t$ captures the structural, content, and temporal features.

\noindent\textbf{Edge-level Training objective.}
Until now, we have established the hidden states of nodes $\bm{\mathcal{H}}_t$ at time window $t$. 
For each edge $(i,j,w)\in \mathcal{E}^t$ with weight $w$, we retrieve the embeddings for the $i$-th and $j$-th node in $\bm{\mathcal{H}}_t$. 
This allows us to calculate its anomalous score as follows:
\begin{equation}
    f(i,j,w)=w\cdot \sigma( \mathbf{W}_{1}\mathbf{h}_i + \mathbf{W}_{2}\mathbf{h}_j-\mu)
\end{equation}
where $\mathbf{h}_i$ and $\mathbf{h}_j$ are the hidden states of the $i$-th and $j$-th node respectively, and $\sigma(\cdot)$ is the sigmoid function. $\mathbf{W}_{1}$ and $\mathbf{W}_{2}$ are the weights in two fully-connected layers.
$\mu$ is a hyperparameter in the score function.
Note that this single layer network can be replaced by more complex networks.

To overcome the scarcity of anomaly data during training, we build a model to optimize one-class (normal) data instead.
In essence, this means that all edges are considered normal during training. 
Inspired by the sampling method proposed in~\cite{KGETH}, we apply a Bernoulli distribution with parameter $\frac{d_i}{d_i+d_j}$ for sampling anomalous edges according to the node degree $d$.
In particular, for each normal edge $(i,j)$ in the graph, we generate an anomalous edge by either replacing node $i$ with node $i'$ (with a probability of $\frac{d_i}{d_i+d_j}$) or replacing node $j$ with node $j'$ (with a probability of $\frac{d_j}{d_i+d_j}$).
Here, $d_i$ and $d_j$ are the degrees of the $i$-th and $j$-th node respectively.
Realizing that the generated edges may still be normal~\cite{DBLP:conf/nips/BordesUGWY13, addgraph}, we propose a margin-based pairwise edge loss in training rather than a strict objective function such as cross entropy, to distinguish between existing edges and generated edges:
\begin{multline}
\label{eq:edge_loss}
    \mathcal{L}_e = \sum_{t=1,...,N} \min \sum_{(i,j,w)\in\mathcal{E}^t} \sum_{(i',j',w)\notin \mathcal{E}^t} \\
    \max \left \{ 0, \gamma + f(i,j,w)-f(i',j',w) \right \} 
\end{multline}
where $\gamma\in (0,1)$ is the margin between the likelihood of normal and anomalous edges, and $f(\cdot,\cdot,\cdot)$ is the aforementioned anomalous edge score function.
Minimizing the loss function $\mathcal{L}_e$ results in a smaller $f(i,j,w)$ and a larger $f(i',j',w)$, thereby achieving our one-class optimization goal.

To enhance efficiency, we aim to select edges of high significance for training.
Specifically, for each pair of normal edge $(i,j,w)$ and negatively sampled edge $(i',j',w)$, we discard it if $f(i,j,w) > f(i',j',w)$ and retain it otherwise, for pair-wise optimization.
The intuition behind is that some edges in snapshots may not be entirely normal after training, and we aim to increase the reliability of normal edges that are used to learn graph representations.
This selective negative sampling paradigm bolsters the stability of \model{} in training.

\noindent\textbf{Multi-granularity Learning.}
Besides the margin loss that differentiates normal and anomalous edges, we introduce an ad-hoc heuristic to form a ``soft-margin" decision boundary.
This means we select graph representations whose distance to a center ranks at specific percentile as the decision boundary's radius~\cite{logbert}.
To this end, we first formulate the graph representation for $\mathcal{G}_t$ by maxpooling its node representations:
\begin{equation}
    \label{eq:graph embedding}
    \bm{\mathcal{R}}_{t} = \text{maxpooling}(\bm{\mathcal{H}}_t)
\end{equation}

At the graph-level, anomalous graphs can be detected via one-class classification training.
The objective $\mathcal{L}_g$ is to learn a minimized hypersphere that encloses graph representations:
\begin{equation}
    \label{eq:sphere learning}
\begin{gathered}
    \min_{R,\mathbf{c},\varepsilon} R^2+C\sum_{t=1}^N \varepsilon_t\\
    s.t.\; || \bm{\mathcal{R}}_t - \mathbf{c}||^2 \leq R^2+\varepsilon_t, \varepsilon_t \geq 0,\; \forall t
\end{gathered}
\end{equation}
where $\mathbf{c}$ and $R$ are the center and radius of the hypersphere respectively, $|| \bm{\mathcal{R}}_t - c||^2$ is the distance between a graph representation and the center, $\varepsilon_t$ is a slack variable introduced for $\bm{\mathcal{R}}_t$ to accommodate outliers during training, and $C$ is a hyperparameter that balance the trade-off between the errors $\varepsilon_t$ and the volume of the sphere.
The objective defined in Eq.~\ref{eq:sphere learning} aims to cluster all training samples within a minimum hypersphere using Lagrange multipliers, similar to SVDD~\cite{DeepSVDD}.
We propose a multi-granularity loss function that considers both edge-level and graph-level objectives:
\begin{equation}
    \mathcal{L} = \mathcal{L}_e + \alpha \mathcal{L}_g + \frac{\lambda}{2} \sum(||\mathbf{W}_{g}||_{2}^2 + ||\mathbf{W}_{a}||_{2}^2 + ||\mathbf{W}_1||_{2}^2 + ||\mathbf{W}_2||_{2}^2)
\end{equation}
where $\mathbf{W}_{a}$ denotes the weights of temporal-attentive transformers.
Hyperparameter $\alpha$ controls the trade-off between edge-level and graph-level violations, and $\lambda$ modulates the weight decay L2 regularizer to avoid overfitting.

\section{Experiments}

\subsection{Experimental Settings}

We evaluate our method in both the new anomalous relation detection setting and the traditional setting: 
(1) Edge-level detection: it aims at detecting anomalous relations in a log sequence, which are the edges in a log graph for a given time window.
For each dataset, we label edges connected to the annotated anomalous logs as anomalies under this new setting.
(2) Interval-level detection: it aims at detecting anomaly time windows which contains anomalous logs. We use this setting for a fair comparison with traditional log anomaly detection methods and more recent graph-based anomaly detection methods. In this setting, we treat a time window as anomaly if it contains any labeled anomalous logs.
This is equivalent to the graph-level detection in our context.

\begin{table}[]
    \centering
    \caption{Statistics of the three datasets.}
    \resizebox{0.4\textwidth}{!}{
    \begin{tabular}{l|ccc}
    \hline
          & BGL & AIT & Sock Shop \\
         \hline
        \# Log Messages & 4,713,494 & 1,074,902 & 14,674 \\
        \# Anomalies & 348,460 & 45,651 & 408 \\
        \# Nodes & 4,393,108 & 1,663,188 & 10,340 \\
        Avg. degree & 11.80 & 15.85 & 13.84 \\
        \# Edges & 25,918,022 & 13,180,752 & 71,540 \\
        \# Anomalous edges & 1,572,696 & 567,906 & 2,468 \\
        \# Graphs & 36,169 & 15,464 & 270 \\
        \# Anomalous graphs & 2,659 & 1,078 & 16 \\
        \hline
    \end{tabular}}
    \label{tab:datasets}
    \vspace{-0.3cm}
\end{table}

\noindent\textbf{Datasets.}
Among several potential candidates, we choose three publicly available datasets or platforms that have been examined by previous researches.
We collect log sequences from these data sources to evaluate the effectiveness of our approach. 
Below we describe the details of the three datasets, and their statistics is shown in Table~\ref{tab:datasets}.

\begin{itemize}
    \item BlueGene/L (BGL)~\cite{BGL}. BGL is an open dataset of logs collected from a BlueGene/L supercomputer system with 131,072 processors and 32,768GB memory. 
    The logs can be categorized into alert (anomalous) and non-alert (normal) messages identified by alert category tags.  

    \item Austrian Institute of Technology (AIT)~\cite{AIT}. 
    AIT (v1.1) is collected from four independent testbeds. Each of the web servers runs Debian and a set of installed services such as Apache2, PHP7, Exim4, Horde, and Suricata. 
    Furthermore, the data includes logs from 11 Ubuntu hosts on which user behaviors were simulated. 

    \item Sock Shop Microservices~\cite{SockShop2017}. 
    Sock Shop is a test bed that can be used to illustrate microservices architectures, demonstrate platforms at talks and meetups, or as a training and education tool. 
    Specifically, we deploy and generate anomalous relations by adding shopping items that have not been browsed by each customer or introducing a large number of items in certain time periods. 
\end{itemize}

\noindent\textbf{Baselines.}
We compare the performance of \model{} with a wide range of baselines.
For the edge-level setting, we consider five graph-based anomaly detection baselines.
For fairness, except for AddGraph~\cite{addgraph} that directly identifies anomalous edges, we use their reconstructed node feature vectors to compute edge scores and evaluate their edge-level performance.

\begin{itemize}

    \item DOMINANT~\cite{dominant}. DOMINANT contains a GCN encoder, a structure reconstruction decoder and a attribute reconstruction decoder. It learns a weighted reconstruction errors as the node anomalous score.

    \item CONAD~\cite{conad}. CONAD first generates augmented graphs based on prior human knowledge of anomaly types, then applies a Siamese GNN to detect node anomalies.  

    \item AnomalyDAE~\cite{anomalydae}. AnomalyDAE uses a structure encoder-decoder to learn structure reconstruction errors and an attribute encoder-decoder to learn feature reconstruction errors. These errors are balanced to form an anomalous score of each node.

    \item MLPAE~\cite{MLPAE}. MLPAE applies a Multi-Layer Perceptron (MLP) autoencoder to detect anomalous nodes without considering structure information in a graph.

    \item AddGraph~\cite{addgraph}. AddGraph incorporates a temporal-attentive RNN into a GCN encoder to learn structure and attribute representations in dynamic graphs. It learns edge scores according to pairwise node latent vectors and detects anomalous edges.

\end{itemize}

\begin{table*}[]
    \centering
    \caption{Edge-level performance (\%) of \model{} and baseline methods.
    \textbf{Bold} numbers denote the best metric among all the methods.
    We ran each model 5 times to get the average results.
    }
    \resizebox{0.98\textwidth}{!}{
    \begin{tabular}{l|ccccc|ccccc|ccccc}
    \hline
        \multirow{2}{*}{Method} & \multicolumn{5}{c|}{BGL} & 
        \multicolumn{5}{c|}{AIT} &
        \multicolumn{5}{c}{Sock Shop} \\
        \cline{2-16}
         & Precision & Recall & F-1 & AUC & AUPR & Precision & Recall & F-1 & AUC & AUPR & Precision & Recall & F-1 & AUC & AUPR \\
        \hline 
        DOMINANT & 35.09 & 90.68 & 50.60 & 42.99 & 24.30 & 39.94 & 89.56 & 55.24 & 42.01 & 43.39 & 11.59 & \textbf{95.63} & 20.68 & 39.83 & 13.33 \\
        CONAD & 32.19 & \textbf{97.83} & 48.44 & 44.84 & 25.99 & 39.93 & 89.54 & 55.23 & 44.48 & 45.47 & 16.31 & 88.92 & 27.56 & 43.82 & 17.49 \\
        AnomalyDAE & 36.34 & 88.27 & 51.48 & 45.31 & 26.03 & \textbf{55.12} & 70.63 & 61.92 & 45.49 & 45.17 & 12.40 & 93.93 & 21.90 & 39.70 & 12.53 \\
        MLPAE & 35.53 & 82.41 & 49.65 & 43.76 & 24.65 & 39.94 & 89.58 & 55.25 & 43.71 & 43.60 & 11.33 & 93.87 & 20.22 & 38.22 & 11.34 \\
        AddGraph & \textbf{48.21} & 66.27 & 55.82 & 54.29 & 33.97 & 48.96 & 85.92 & 62.38 & 46.16 & 46.94 & 34.49 & 84.51 & 48.99 & 51.39 & 58.52 \\
        \hline
        $\model{}^{\xi}$ & 38.94 & 74.26 & 51.09 & 50.73 & 30.06 & 44.91 & 89.87 & 59.89 & 45.16 & 46.30 & 35.79 & 80.14 & 49.48 & 50.74 & 52.88 \\
        $\model{}^{\wr}$ & 40.60 & 73.31 & 52.26 & 50.34 & 30.82 & 45.17 & 90.71 & 60.31 & 45.83 & 46.12 & 32.28 & 82.77 & 46.45 & 48.57 & 52.39 \\
        $\model{}^{\dagger}$ & 39.53 & 89.56 & 54.85 & 52.97 & 31.07 & 50.08 & \textbf{93.30} & 65.18 & 46.29 & 46.53 & 50.86 & 82.76 & 63.01 & 61.85 & 65.42 \\
        \model{} & 47.09 & 86.06 & \textbf{60.87} & \textbf{56.56} & \textbf{38.99} & 54.15 & 90.81 & \textbf{67.84} & \textbf{49.09} & \textbf{48.66} & \textbf{56.02} & 91.00 & \textbf{69.35} & \textbf{61.93} & \textbf{68.37} \\
        \hline
        \end{tabular}}
    \label{tab:overall_res}
    \vspace{-0.3cm}
\end{table*}

For the interval-level setting, existing works focusing on logs can be divided into two categories: 1) sequence-based methods, including traditional methods such as PCA~\cite{large-scale-mining-console-logs}, Isolation Forest~\cite{isolation-forest-2008}, OCSVM~\cite{high-dimension-distribution} and deep learning-based methods such as DeepLog~\cite{ccs-2017-deeplog}, LogAnomaly~\cite{unsupervised-sequential}, LogBERT~\cite{logbert};
and 2) graph-based methods, including LogGD~\cite{loggd}, LogFlash~\cite{logflash}, and DeepTraLog~\cite{DeepTraLog}.

\begin{itemize}
    \item Principal Component Analysis~(PCA)~\cite{large-scale-mining-console-logs}. PCA builds a counting matrix according to the log event frequency and then maps the matrix into a latent space to detect anomalous sequences.

    \item Isolation Forest~(iForest)~\cite{isolation-forest-2008}. An unsupervised learning method that represents features as tree structures for anomaly detection.

    \item One-class SVM~(OCSVM)~\cite{high-dimension-distribution}. A well-known one-class classification method by building a feature matrix based on the norm data for anomaly detection.

    \item DeepLog~\cite{ccs-2017-deeplog}. DeepLog uses LSTM to capture patterns of normal log sequences and further identifies anomalous log sequences based on log key predictions.

    \item LogAnomaly~\cite{unsupervised-sequential}. LogAnomaly proposes template2vec to extract log template semantics and use LSTM to detect sequential and quantitative log anomalies. 

    \item LogBERT~\cite{logbert}. LogBERT uses BERT to encode each log sequence into a feature space by self-supervision, and detect anomalous log sequences via hypersphere learning.

    \item LogGD~\cite{loggd}. LogGD constructs directed graph by connecting log templates following sequential relations, and identifies anomalies via graph classification based on Graph Transformer network.

    \item LogFlash~\cite{logflash}. LogFlash builds a time-weighted control flow graph (TCFG), where nodes are log templates and edges represent the transition between them, and compare log streams with TCFG to find deviations.

    \item DeepTraLog~\cite{DeepTraLog}. DeepTraLog constructs trace event graph (TEG) to represent various relations between the span/log events of the trace. It learns a gated GNN-based SVDD representation for each TEG and identifies anomalies via hypersphere learning.
\end{itemize}

To evaluate the impact of individual components in \model{} on the final performance, we also conduct experiments on different \model{} variants: 
\begin{itemize}
    \item $\model{}^{\xi}$. 
    In this variant, \model{} applies a rule-based (instead of prompt-based) field extraction during graph configuration. This allows us to evaluate the significance of accurate identification of system entities and their interrelations with log events.
    
    \item $\model{}^{\wr}$.
    This version of \model{} is designed without the use of transformer encoder, removing its ability to capture temporal features. The purpose is to investigate the significance of temporal features.
    
    \item $\model{}^{\dagger}$.
    This variant removes multi-granularity learning from the training process of \model{}, thereby examining the importance of global features in detecting anomalies.
    
\end{itemize}

\noindent\textbf{Metrics.}
We measure the model performance on anomaly detection based on three widely-used classification metrics, including Precision, Recall, and F-1 score, as well as two ranking metrics, including AUC and AUPR score. 


\noindent\textbf{Implementation Details.}
All GNN models in our research are built on PyTorch Geometric (PyG) framework. 
These models are configured with two layers, with input channels set at 768, and output channels at 1,024. 
For Sentence-BERT and BART, we use their pre-trained models, namely \textit{bert-base-uncased} and \textit{facebook/bart-base}, from Hugging Face.
For field extraction, we either fine-tune BART over 100 epochs using 10-shot training samples or use pre-defined regular expressions.
For anomaly detection, we split the log sequences into a ratio of 6:1:3, where 60\% as training set, 10\% as validation set, and 30\% as test set.
We apply an unsupervised learning paradigm~\cite{high-dimension-distribution, unsupervised-sequential, DeepTraLog} where only normal log sequences are used for training, and train each model for 100 epochs.
Hyperparameters are adjusted via grid search on the validation set.
Specifically, we use AdamW optimizer~\cite{adamW} with a learning rate of 1e-3, $\mu$ of 0.3, $\gamma$ of 0.5, global weight $\alpha$ of 1, and weight decay $\lambda$ of 5e-7.
Our analysis operates with a window size of 60 seconds.
Our work is conducted in a leading industry company using an NVIDIA RTX A4500 GPU. 
Our \model{} has been deployed to monitor internal cloud system log data for anomaly detection.

\subsection{Experimental Results}

\noindent\textbf{Edge-level Performance.}
We first compare \model{} with baseline methods in terms of their edge-level performance.
As shown in Table~\ref{tab:overall_res}, we observe that:
(1) \model{} outperforms all baseline methods in F-1 score, AUC score and AUPR score.
This demonstrates the efficacy of our approach in identifying anomalous relations between log fields and log events.
(2) While some baseline methods like DOMINANT, CONAD, AddGraph in BGL dataset, and AnomalyDAE, MLPAE in Sock Shop dataset, achieve high recall scores, and AnomalyDAE in AIT dataset, AddGraph in BGL dataset achieve high precision scores, their F-1 scores are relatively low. 
This suggests that they either adopt an overly cautious stance towards anomalies or produce a high number of false positives by erroneously classifying many samples as anomalies.
(3) Edge-level anomaly detection is notably more challenging compared to interval-level anomaly detection, e.g., no method in the three datasets achieved a precision score exceeding 60\% or an F-1 score above 70\%.
(4) Those method optimized on edge-level, i.e., AddGraph and \model{}, achieve better precision, F-1, AUC and AUPR scores across all datasets. 
Specifically, they exhibit larger advantages over other methods in our generated Sock Shop datasets that contain specific anomalous relations, substantiating our hypothesis that edge-level learning can better detect anomalous relations that are elusive to other methods.  
(5) While some methods, such as AnomalyDAE, excel in one dataset, achieving a 61.92\% F-1 score in the AIT dataset, they flounder in others, dropping to a 21.90\% F-1 score in the Sock Shop dataset, for instance.
(6) Compared to AddGraph, \model{} achieves superior performance, especially recall scores, across all three datasets. 
This demonstrates the advantages of our graph configuration and graph-based edge-level anomaly detection method.

\begin{table}[]
    \centering
    \caption{Interval-level performance (\%) in the BGL dataset. We ran each model 5 times to get the average results.}
    \resizebox{0.4\textwidth}{!}{
    \begin{tabular}{l|ccccc}
    \hline
        Method & Precision & Recall & F-1 & AUC & AUPR \\
        \hline
        PCA & 9.04 & \textbf{98.12} & 16.56 & 55.64 & 9.03 \\
        iForest & \textbf{100.00} & 14.74 & 25.70 & 57.37 & 21.64 \\ 
        OCSVM & 1.09 & 12.48 & 2.00 & 28.22 & 7.32 \\
        \hline
        DeepLog & 89.02 & 80.54 & 84.57 & 89.26 & 70.17 \\
        LogAnomaly & 91.40 & 79.32 & 84.93 & 92.98 & 75.21  \\ 
        LogBERT & 91.47 & 92.69 & 92.07 & 96.33 & 82.70  \\
        \hline
        LogGD & 90.89 & 93.31 & 92.08 & 96.91 & 81.74  \\
        LogFlash & 82.46 & 86.73 & 84.54 & 86.78 & 74.52 \\ 
        DeepTraLog & 79.48 & 97.68 & 87.64 & 84.77 & 70.92  \\ 
        \hline
        $\model{}^{\xi}$ & 88.35 & 89.86 & 89.10 & 95.63 & 80.44  \\
        $\model{}^{\wr}$ & 89.24 & 90.18 & 89.51 & 96.07 & 79.91  \\
        $\model{}^{\dagger}$ & 89.73 & 91.64 & 90.67 & 96.31 & 81.65 \\
        \model{} & 90.82 & 94.57 & \textbf{92.66} & \textbf{98.18} & \textbf{84.69} \\
        \hline
    \end{tabular}}
    \label{tab:graph_level_bgl}
    \vspace{-0.4cm}
\end{table}

\noindent\textbf{Interval-level Performance.}
We further evaluate \model{} on the common interval-level protocol to demonstrate its effectiveness.
Due to space limit, we only present the results of the widely used BGL dataset in Table~\ref{tab:graph_level_bgl}. We observe that:
(1) Compared to edge-level detection results, \model{} achieves much higher Precision (and F-1).
Significantly, \model{} surpasses all methods with a leading F-1 of 92.66\%, AUC of 98.18\%, and AUPR of 84.69\%. 
This demonstrates the proficiency of \model{} in capturing conventional anomalies in addition to relational ones.
(2) Traditional sequence-based methods such as PCA, iForest, and OCSVM have significantly lower performance across all metrics. 
For instance, PCA suffers severely in Precision (9.04\%), 
while iForest has a notably poor Recall (14.74\%). 
OCSVM has the lowest performance among the three, with a negligible F-1 score of 2.00\%.
(3) DL sequence-based methods significantly outperform the traditional ones.
Among these, LogBERT outperforms DeepLog and LogAnomaly with a F-1 score of 92.07\% and an AUC of 96.33\%. 
This showcases the effectiveness of transformers in capturing both long-term dependencies and semantic relations in log sequences.
(4) Among graph-based methods, LogGD shows competitive results with an F-1 score of 92.08\%, even slightly better than LogBERT. 
This suggests that the inclusion of both transformers and graph structures in hypersphere learning could benefit anomaly detection. 

\begin{figure}
    \centering
    \includegraphics[width=0.35\textwidth]{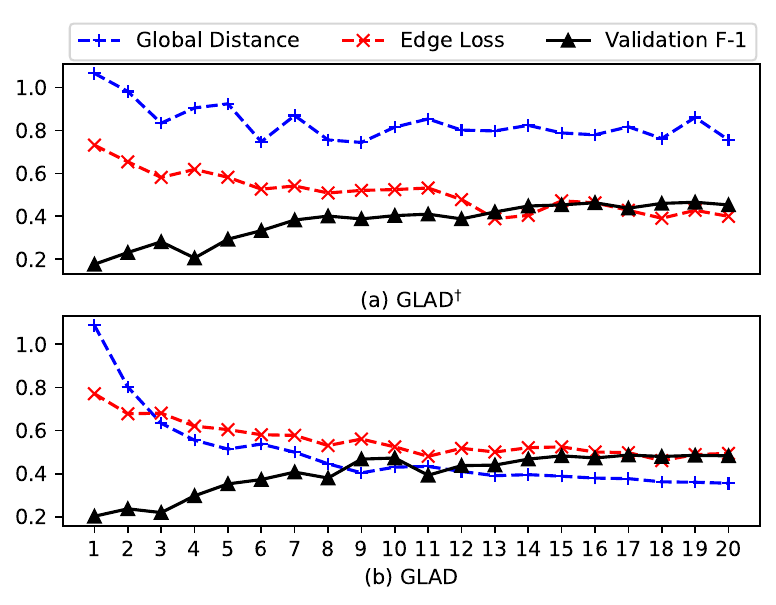}
    \caption{Ablation study of multi-granularity learning.
    }
    \label{fig:ablation}
    \vspace{-0.3cm}
\end{figure}

\begin{figure}[]
    \centering        \includegraphics[width=0.35\textwidth]{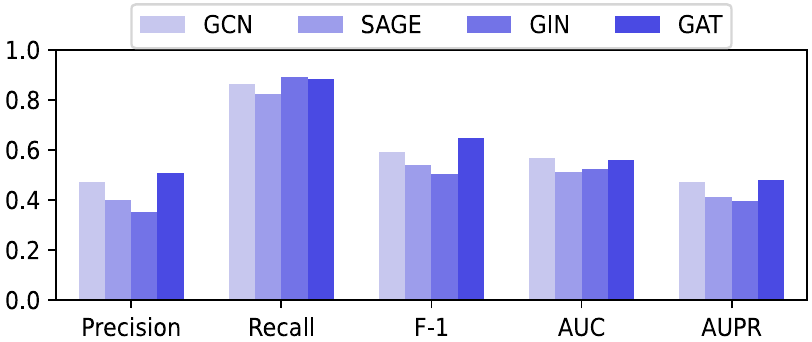}
    \caption{\model{} performance using different GNN encoders.
    }
    \label{fig:encoder_study}
    \vspace{-0.3cm}
\end{figure}

\noindent\textbf{Ablation Study.}
With results of \model{} variants shown in Table~\ref{tab:overall_res} and Table~\ref{tab:graph_level_bgl}, we observe that: 
(1) The performance gap between $\model{}^{\xi}$ and \model{}---roughly 9\% F-1 improvement for edge-level detection in BGL---reveals the benefit of employing prompt-based filed extraction in graph configuration, thereby enhancing the effectiveness of \model{} in detecting anomalies.
(2) The difference between $\model{}^{\wr}$ and \model{} underscores the significant role of temporal-attentive transformers.
With the incorporation of temporal features, \model{} gains over 8\% (and 3\%) F-1 increases when detecting anomalous relations (and intervals) in BGL dataset.
However, $\model{}^{\wr}$ still outperforms numerous baseline methods, asserting the robustness of our graph-based framework.
(3) The comparison between $\model{}^{\dagger}$ and \model{} indicates the positive impact of incorporating global features in anomaly detection, as evidenced by the superior F-1, AUC and AUPR scores of \model{}.

To further illustrate how multi-granularity learning benefits \model{} during training, we record the normalized global distance ($\mathcal{L}_g$ in Eq.~\ref{eq:sphere learning}), edge loss ($\mathcal{L}_e$ in Eq.~\ref{eq:edge_loss}) and validation F-1 scores after each training epoch in BGL dataset. 
In Figure~\ref{fig:ablation},
we observe that:
(1) The comparison between $\model{}^{\dagger}$ and \model{} in terms of global distance shows that hypersphere learning effectively clusters normal samples in the graph embedding space, i.e., the converged normalized global distance of \model{} is less than half of that of $\model{}^{\dagger}$.
(2) The comparison between $\model{}^{\dagger}$ and \model{} in terms of edge loss and validation F-1 score shows that hypersphere learning further improves the anomaly detection performance, i.e., the edge loss of \model{} decreases more stably and its validation F-1 outperforms that of $\model{}^{\dagger}$ in the later training stage.

We also analyze the impact of using different GNN encoders, i.e., GCN~\cite{GCN}, SAGE~\cite{SAGE}, GIN~\cite{GIN}, GAT~\cite{GAT}, on \model{}'s performance in the BGL dataset.
Figure~\ref{fig:encoder_study} reveals that the performance of \model{} remains consistently robust across diverse GNN encoders, though some models excel in specific evaluation metrics, e.g., GCN and GIN show superior performance in terms of F-1, AUC, and AUPR scores. 
This resilience against changes in DL models demonstrates that \model{} can be flexibly deployed using various combinations of state-of-the-art architectures.

\begin{table}[]
\caption{Two proposed prompts for field extraction.}
\centering
\resizebox{0.35\textwidth}{!}{
\begin{tabular}{l}
    \hline
    Prompt $\mathbf{P}_1$ \\
    \hline
    $\mathbf{P}^{+}: \left \langle candidate\_span \right \rangle$ is a/an $\left \langle entity\_type \right \rangle$ entity \\
    $\mathbf{P}^{-}: \left \langle candidate\_span \right \rangle$ is not a named entity  \\
    \hline
    Prompt $\mathbf{P}_2$ \\
    \hline
    $\mathbf{P}^{+}: \left \langle entity\_type \right \rangle$ = $\left \langle candidate\_span \right \rangle$ \\
    $\mathbf{P}^{-}: \left \langle candidate\_span \right \rangle$ = none \\
    \hline
\end{tabular}}
\label{tab:seq2seq prompts}
\end{table}

\begin{table}[]
    \centering
    \caption{Performance of rule-based field extraction v.s. prompt-based $n$-shot field extraction.}
    \begin{tabular}{cc|ccc}
        \hline
        \multicolumn{2}{c|}{Technique} & Pre. & Rec. & F-1  \\
        \hline
        \multicolumn{2}{c|}{regex} & 36.48 & 44.28 & 40.00 \\
        \hline
        \multirow{3}{*}{$\mathbf{P}_1$} & 1-shot & 16.53 & 59.34 & 25.86 \\
        & 5-shot & 28.33 & 74.38 & 41.03 \\
        & 10-shot & \textbf{66.28} & 85.22 & \textbf{74.57} \\
        \hline
        \multirow{3}{*}{$\mathbf{P}_2$} & 1-shot & 17.89 & 58.14 & 27.36 \\
        & 5-shot & 28.00 & 73.76 & 40.59 \\
        & 10-shot & 64.68 & \textbf{87.82} & 74.49 \\
        \hline
    \end{tabular}
    \label{tab:field extraction res}
    \vspace{-0.3cm}
\end{table}

\noindent\textbf{Field Extraction.}
To investigate the effectiveness of our field extraction method, we annotate $n$ log messages with two prompts ($\mathbf{P}_1$, $\mathbf{P}_2$ in Table~\ref{tab:seq2seq prompts}) for each field type and train corresponding field extraction models.
Note that we use $\mathbf{P}_1$ in \model{} for graph construction due to slightly better anomaly detection performance.
As shown in Table~\ref{tab:field extraction res}, with only 5-shot learning, our field extraction model is as competitive as hand-crafted rules in terms of F-1 scores.
Our method significantly outperforms rule-based method when conducting 10-shot learning, which explains the superiority of \model{} over $\model{}^{\xi}$ and suggests the practicability of our few-shot method in low-resource scenarios where annotations are limited.

\begin{figure}
    \centering
    \includegraphics[width=0.35\textwidth]{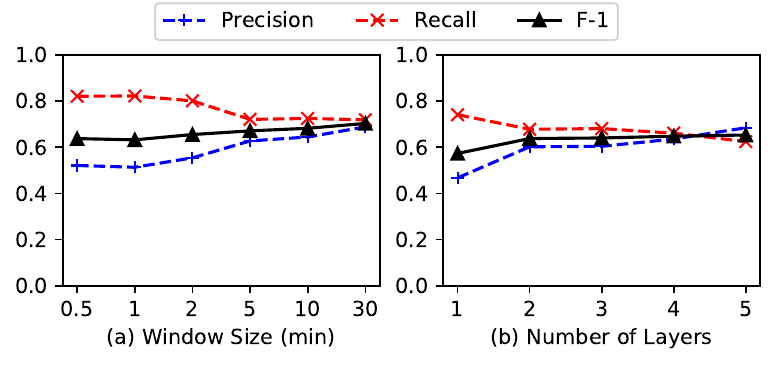}
    \caption{\model{} performance with different parameters.}
    \label{fig:parameter_study}
    \vspace{-0.4cm}
\end{figure}

\noindent\textbf{Parameter Study.}
We investigate the impact of two critical hyperparameters---window size and the number of GNN layers---on \model{}'s performance in the BGL dataset.
In each experiment, one hyperparameter is altered while the rest are held constant.
As shown in Figure~\ref{fig:parameter_study}, the performance of \model{}, especially the F-1 scores, is robust to variations in these hyperparameters. 
This indicates that \model{} maintains its effectiveness across a range of configurations, highlighting its suitability for deployment in real-world scenarios.
Specifically, Figure~\ref{fig:parameter_study}a suggests that a longer monitoring period leads to higher precision but lower recall.
This implies that the window size can be tuned to balance between achieving higher true positive rates and reducing false positive rates.
Similar strategy (Figure~\ref{fig:parameter_study}b) is applicable to configuration of GNN layers.



\begin{table}[ht]
\centering
\caption{Comparison of total training and testing overheads, and average overheads per log for different methods.}
\begin{tabular}{l|cc|cc}
\hline
\multirow{2}{*}{Method} & \multicolumn{2}{c|}{Training Time} & \multicolumn{2}{c}{Testing Time} \\
\cline{2-3} \cline{4-5}
 & Total (s) & Avg. (ms) & Total (s) & Avg. (ms) \\
\hline
PCA & 87.36 & 5 & 0.61 & 0 \\
iForest & 0.00 & 0 & 4.63 & 0 \\
OCSVM & 235.79 & 15 & 107.13 & 8 \\
\hline
DeepLog & 2321.21 & 155 & 1595.18 & 125 \\
LogAnomaly & 4420.02 & 296 & 2625.36 & 196 \\
LogBERT & 1950.31 & 130 & 470.14 & 37 \\
\hline
LogGD & 1753.82 & 117 & 1050.07 & 82 \\
LogFlash & 678.52 & 45 & 286.33 & 22 \\
DeepTraLog & 1261.55 & 84 & 796.81 & 62 \\
\hline
\model{} & 2490.40 & 166 & 1170.73 & 92 \\
\hline
\end{tabular}
\label{tab:efficiency}
\end{table}

\noindent\textbf{Efficiency Analysis.}
We compare the training and testing time of different methods in the BGL dataset. 
As shown in Table~\ref{tab:efficiency}, traditional methods such as PCA, iForest, and OCSVM show small overheads as they are rather simple, among which OCSVM exhibits a rather high overhead due to construction of a feature matrix based on norm data for anomaly detection.
DL sequence-based methods such as DeepLog, LogAnomaly, and LogBERT, generally possess higher overheads due to their complex architectures. 
Among them, LogAnomaly has the highest overhead due to complex template2vec learning process and low parallelism. While graph-based methods show rather lower overheads, underscoring the computational efficiency of graph structures.
Interestingly, \model{} provides a training and testing overhead of 166 and 92 milliseconds per log, which is notably less than that of LogAnomaly and comparable to DeepLog. 
Given the sophisticated transformer and GNN architectures of \model{}, its relatively small overhead underscores our efficient design.
This can be attributed to the direct application of temporal-attentive transformers on graph features, avoiding both tokenization and embedding, thereby increasing parallel computation.

\begin{figure}[]
    \centering
    \includegraphics[width=0.47\textwidth]{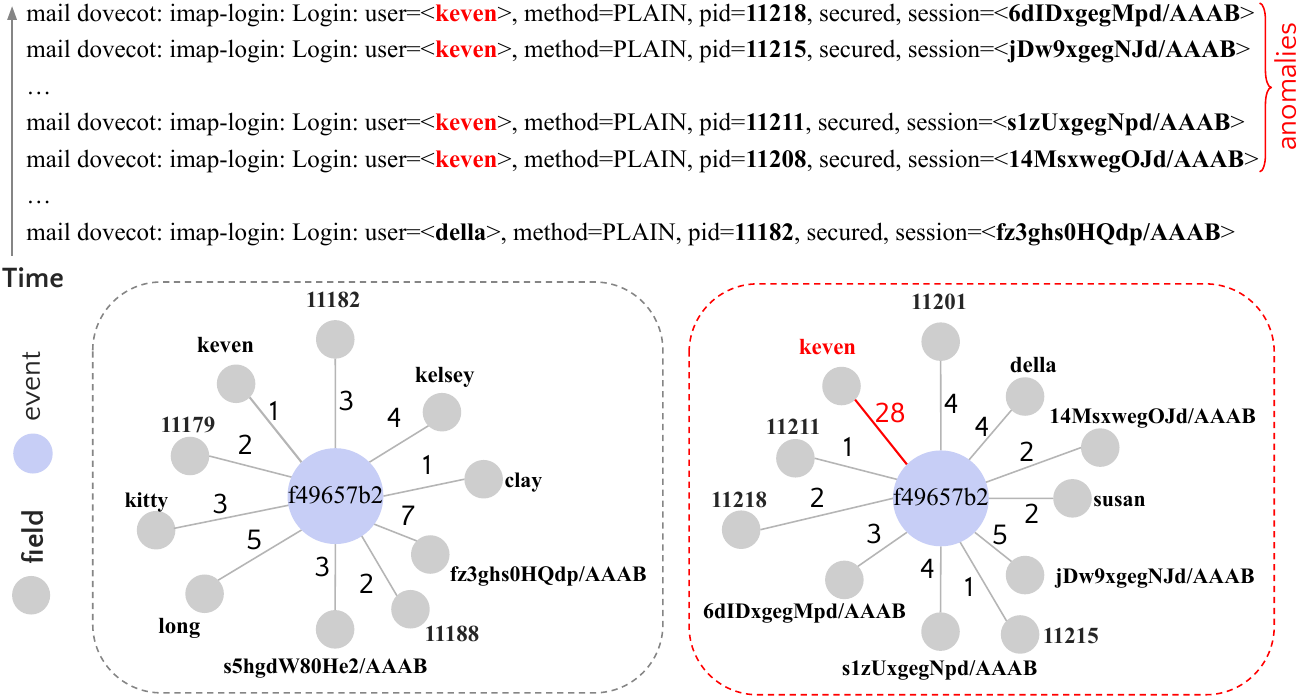}
    \caption{Visualization of a normal graph (left) and anomalous graph (right).
    Edge marked in \textcolor{red}{red} denotes anomalous relation.}
    \label{fig:case_study}
    \vspace{-0.4cm}
\end{figure}

\noindent\textbf{Case Study.}
To provide deeper insight into the performance of our graph-based anomaly detection, we visualize two sample graphs in Figure~\ref{fig:case_study}.
In this case, all log messages share the same event template ``f49657b2'', and the anomalous relation manifests as the user ``keven'' making frequent requests to a server (28 times) compared to other users whose requests are considerably fewer. 
Existing methods that neglect system interactions cannot identify such anomalies, as they do not consider relations among system components.
Our \model{}, however, successfully detects these anomalies by considering both the edge weight values and temporal patterns in a sequence of graphs. 
The comparison of the two constructed graphs also demonstrate the interpretability of our graph-based approach.

\section{Conclusions}
In this paper, we proposed a Graph-based Log Anomaly Detection framework, \model{}, which considers relational patterns in addition to log semantics and sequential patterns for system relation anomaly detection. 
First, a field extraction module utilizing prompt-based few-shot learning is used to extract field information, e.g., \textit{service}, \textit{user}, from log contents. 
Then, with the log events and fields extracted, dynamic log graphs can be constructed for sliding windows with events and fields as nodes, and the relations between them as edges. 
Finally, a temporal-attentive graph edge anomaly detection model is introduced for detecting anomalous relations from the dynamic log graphs, where a GNN-based encoder facilitated with transformers is used to model the structural, content, and temporal features. 
Experiments conducted on three datasets demonstrated the effectiveness of \model{} on system relation anomaly detection using system logs and providing deep insights into the anomalies.

\bibliographystyle{IEEEtran}
\bibliography{IEEEabrv,ref}

\end{document}